\documentstyle[colap,epsf]{article}

\title{\vspace{-0.5in} Clustering Words with the MDL Principle}
\author{Hang Li and Naoki Abe \\ Theory NEC Laboratory,
  RWCP\thanks{Real World Computing Partership} \\ c/o C\&C Research Laboratories, NEC \\ 4-1-1 Miyazaki
  Miyamae-ku, Kawasaki, 216 Japan \\ 
  \{lihang,abe\}@sbl.cl.nec.co.jp } 

\newcommand{\tab}{\hspace{5mm}}

\begin{document}
\maketitle
\vspace{-0.5in}
\begin{abstract}
  We address the problem of automatically constructing a thesaurus
  (hierarchically clustering words) based on corpus data.  We view the
  problem of clustering words as that of estimating a joint
  distribution over the Cartesian product of a partition of a set of
  nouns and a partition of a set of verbs, and propose an estimation
  algorithm using simulated annealing with an energy function based on
  the Minimum Description Length (MDL) Principle.  We empirically
  compared the performance of our method based on the MDL Principle
  against that of one based on the Maximum Likelihood Estimator, and
  found that the former outperforms the latter. We also evaluated the
  method by conducting pp-attachment disambiguation experiments using
  an automatically constructed thesaurus. Our experimental results
  indicate that we can improve accuracy in disambiguation by using
  such a thesaurus.
\end{abstract}

\section{Introduction}
Recently various methods for automatically constructing a thesaurus
(hierarchically clustering words) based on corpus data have been
proposed \cite{Hindle90,Brown92,Pereira93,Tokunaga95}. The realization
of such an automatic construction method would make it possible to a)
save the cost of constructing a thesaurus by hand, b) do away with the
subjectivity inherent in a hand made thesaurus, and c) make it easier
to adapt a natural language processing system to a new domain. 

Although many of the proposed methods have proved to be effective, the
word clustering problem is still a problem which needs further
investigation. In this paper, we propose a new method for automatic
construction of thesauruses. Specifically, we view the problem of
automatically clustering words as that of estimating a joint
distribution over the Cartesian product of a partition of a set of
nouns (in general, any set of words) and a partition of a set of verbs
(in general, any set of words), and propose an estimation algorithm
using simulated annealing with an energy function based on the Minimum
Description Length (MDL) Principle. The MDL Principle is a
well-motivated and theoretically sound principle for data compression
and estimation from information theory and statistics. As a strategy
of statistical estimation, MDL is guaranteed to be near optimal.

We empirically evaluated the effectiveness of our method. In
particular, we compared the performance of our method based on the MDL
Principle against that of one based on the Maximum Likelihood
Estimator (MLE for short). We found that the MDL-based method performs
better than the MLE-based method. We also evaluated our method by
conducting structural (pp-attachment) disambiguation experiments using a
thesaurus automatically constructed by it and found that
disambiguation results can be improved.

Since some words never occur in a corpus, and thus cannot be reliably
classified by a method solely based on corpus data, we propose to
combine the use of an automatically constructed thesaurus and that of
a hand made thesaurus in disambiguation. We conducted some experiments
in order to test the effectiveness of this strategy. Our experimental
results indicate that combining an automatically constructed thesaurus
and a hand made thesaurus widens the `coverage'\footnote{`Coverage'
  refers to the proportion (in percentage) of test data for which the
  disambiguation method can make a decision.} of disambiguation, while
maintaining high `accuracy'\footnote{`Accuracy' refers to the success
  rate, given that the disambiguation method makes a decision.}.

\section{The Problem Setting}
Many of the methods of automatically constructing a thesaurus based on
corpus data consist of the following three steps: (i) Extract
co-occurrence data (e.g., case frame data, adjacency data) from a
corpus, (ii) Starting from a single class (or each word composing its
own class), divide (or merge) word classes based on the co-occurrence
data using some similarity (distance) measure. (The former approach is
called `divisive,' the latter `agglomerative.') (iii) Repeat step (ii)
until some stopping condition is met, to construct a thesaurus tree. 
The method we propose here consists of the same three steps.

Suppose available to us are data like those in Figure~\ref{fig:data},
which are co-occurrence data between verbs and their objects,
extracted from a corpus (step (i)). We then view the problem of
clustering words as that of estimating a probabilistic model
(representing a probability distribution) that gave rise to such data.
We define the probabilistic model in the following way.  We first
define a noun partition $\cal P_N$ over a given set of nouns $\cal N$
and a verb partion $\cal P_V$ over a given set of verbs $\cal V$. A
noun partition is any set $\cal P_N$ satisfying ${\cal P_N} \subseteq
2^{\cal N}$, $\cup_{C_i \in {\cal P_N}} C_i = {\cal N}$ and $\forall
C_i, C_j \in {\cal P_N}, C_i\cap C_j =\emptyset$. A verb partition
${\cal P_V}$ is defined analogously. In this paper, we call a member
of a noun partition a `noun cluster,' and a member of a verb partition
a `verb cluster.' We refer to a member of the Cartesian product of a
noun partition and a verb partition ( $\in {\cal P_N} \times {\cal
  P_V}$ ) simply as a `cluster.' We then define a probabilistic model
(or a joint distribution), written $P(C_n,C_v)$, where random variable
$C_n$ assumes a value from a {\em fixed} noun partition $\cal P_N$,
and $C_v$ a value from a {\em fixed} verb partition $\cal P_V$. Within
a given cluster, we assume that each element is generated with equal
probability, i.e.,
\begin{equation}\label{eq:basic} \forall n \in C_n, \forall v \in C_v,
 P(n,v) = \frac{P(C_n,C_v)}{|C_n \times C_v|}. \end{equation}
 Figure~\ref{fig:model} exhibits two example models which might have
 given rise to the data in Figure~\ref{fig:data}.

In this paper, we assume that the observed data are generated by a
model belonging to the class of models just described, and select a
model which best explains the data. As a result of this, we obtain
both noun clusters and verb clusters. This problem setting is based on
the intuitive assumption that similar words occur in the same context
with roughly equal likelihood, as is made explicit in equation
(\ref{eq:basic}). Thus selecting a model which best explains the given
data is equivalent to finding the most appropriate classification of
words based on their co-occurrence.
\setlength{\unitlength}{1mm}
\begin{figure*}[htb]
\begin{center}
\begin{picture}(70,70)(0,0)
\put(10,10){rice}
\put(10,20){bread}
\put(10,30){beer}
\put(10,40){wine}
\put(10,50){OBJ}
\put(30,10){4}
\put(30,20){4}
\put(30,30){0}
\put(30,40){0}
\put(28,50){eat}
\put(40,10){0}
\put(40,20){0}
\put(40,30){5}
\put(40,40){3}
\put(36,50){drink}
\put(50,10){0}
\put(50,20){2}
\put(50,30){1}
\put(50,40){1}
\put(46,50){make}
\put(25,5){\line(1,0){30}}
\put(25,5){\line(0,1){40}}
\put(25,45){\line(1,0){30}}
\put(55,45){\line(0,-1){40}}
\end{picture}
\caption{An example of co-occurrence data}
\label{fig:data}
\end{center}
\end{figure*}

\begin{figure*}[htb]
\begin{center}
\begin{tabular}{cc}
\begin{picture}(70,70)(0,0)
\put(10,10){rice}
\put(10,20){bread}
\put(10,30){beer}
\put(10,40){wine}
\put(10,50){OBJ}
\put(35,5){\line(0,1){40}}
\put(45,5){\line(0,1){40}}
\put(25,25){\line(1,0){30}}
\put(28,50){eat}
\put(36,50){drink}
\put(46,50){make}
\put(28,15){$0.4$}
\put(38,15){$0$}
\put(48,15){$0.1$}
\put(28,35){$0$}
\put(38,35){$0.4$}
\put(48,35){$0.1$}
\put(25,5){\line(1,0){30}}
\put(25,5){\line(0,1){40}}
\put(25,45){\line(1,0){30}}
\put(55,45){\line(0,-1){40}}
\end{picture} &
\begin{picture}(70,70)(0,0)
\put(10,10){rice}
\put(10,20){bread}
\put(10,30){beer}
\put(10,40){wine}
\put(10,50){OBJ}
\put(35,5){\line(0,1){40}}
\put(45,5){\line(0,1){40}}
\put(28,50){eat}
\put(36,50){drink}
\put(46,50){make}
\put(28,25){$0.4$}
\put(38,25){$0.4$}
\put(48,25){$0.2$}
\put(25,5){\line(1,0){30}}
\put(25,5){\line(0,1){40}}
\put(25,45){\line(1,0){30}}
\put(55,45){\line(0,-1){40}}
\end{picture} \\
Model1 & Model2 \\
\end{tabular}
\caption{Two example models}
\label{fig:model}
\end{center}
\end{figure*}

\section{Clustering with MDL}
We now turn to the question of what strategy (or criterion) we should
employ in order to estimate the best model. Our choice is the Minimum
Description Length (MDL) Principle
\cite{Rissanen78,Rissanen83,Rissanen84,Rissanen86,Rissanen89}, a
well-known principle of data compression and estimation from
information theory and statistics. MDL stipulates that the best
probability model for given data is that model which requires the
least code length for encoding the model itself and the given data
relative to it.\footnote{We refer the interested reader to \cite{Li95}
  for explanation of rationals behind using MDL in natural language
  processing.} We refer to the code length for the model as the `model
description length' and that for the data the `data description
length.'

We apply MDL to the problem of estimating a model consisting of a pair
of partitions as described above. In this context, a model with less
clusters, such as Model 2 in Figure~\ref{fig:model}, tends to be
simpler (in terms of the number of parameters), but also tends to have
a poorer fit to the data. In contrast, a model with more clusters,
such as Model 1 in Figure~\ref{fig:model}, is more complex, but tends
to have a better fit to the data. Thus, there is a trade-off
relationship between the simplicity of a model and the goodness of fit
to the data. The model description length quantifies the simplicity
(complexity) of a model, and the data description length quantifies
the goodness of fit to the data. According to MDL, the model which
minimizes the sum total of the two types of description lengths should
be selected.

In what follows, we will describe in detail how to calculate the
description length in our current context, as well as our simulated
annealing algorithm based on MDL.

\subsection{Calculating Description Length}
We will now describe how to calculate the description length for a
model. Recall that each model is specified by the Cartesian product of
a noun partition and a verb partition, and a number of parameters. 
Here we let $k_n$ denote the size of the noun partition, and $k_v$ the
size of the verb partition. Then, there are $k_n \cdot k_v - 1$ free
parameters in a model.

Given a model $M$ and data $S$, its total description length
$L(M)$\footnote{$L(M)$ depends on $S$, but we will leave $S$
  implicit.} is computed as the sum of the model description length
$L_{mod}(M)$, the parameter description length $L_{par}(M)$, and the
data description length $L_{dat}(M)$ (we also sometimes refer to
$L_{mod}(M) + L_{par}(M)$ as the model description length), namely,
\begin{equation} L(M) = L_{mod}(M) + L_{par}(M) + L_{dat}(M).
\end{equation}

We employ the `binary noun clustering method,' in which $k_v$ is fixed
at $|{\cal V}|$ and we are to decide whether $k_n=1$ or $k_n=2$. This
is as if we view the nouns as entities and the verbs as features and
classify the entities based on their features. Since there are
$2^{|{\cal N}|}$ subsets of the set of nouns ${\cal N}$, and for each
{\em binary} noun partition we have two different subsets (a special
case of which is when one subset is ${\cal N}$ and the other the empty
set $\emptyset$), the number of possible binary noun partitions is
$2^{|{\cal N}|}/2= 2^{|{\cal N}|-1}$. Thus for each binary noun
partition we need $- \log \frac{1}{2^{(|{\cal N}|-1)}}= |{\cal N}|-1$
bits\footnote{Throughout the paper `$\log$' denotes the logarithm to
  the base 2.} to describe it.\footnote{For further explanation, see
  for example \cite{Quinlan89}.} Hence $L_{mod}(M)$ is calculated
as\footnote{The exact formulation of $L_{mod}(M)$ is subjective, and
  it depends on the coding scheme used for the description of the
  models.}
\begin{equation} L_{mod}(M) = |{\cal N}| - 1. \end{equation}
  $L_{par}(M)$ is calculated by \begin{equation} L_{par}(M) =
  \frac{k_n \cdot k_v - 1}{2} \cdot \log |S|, \end{equation} where
  $|S|$ denotes the data size, and $k_n \cdot k_v - 1$ is the number
  of free parameters in the model. As is well known, it is best to use
  $-\log \frac{1}{\sqrt{|S|}}$ $= \frac{1}{2}\cdot \log |S|$ bits to
  describe each of the parameters, since the standard deviation of the
  maximum likelihood estimation of each parameter is of order
  $\frac{1}{\sqrt{|S|}}$, and hence describing each parameter using
  more than $O(\frac{1}{2} \cdot \log |S|)$ bits would be
  wasteful for the estimation accuracy possible with the given data
  size. Finally, $L_{dat}(M)$ is calculated by
 \begin{equation} L_{dat}(M) = - \sum_{(n,v)\in {\cal N} \times {\cal V}}
 f(n,v) \cdot \log\hat{P}(n,v),\end{equation} where $f(n,v)$ denotes
 the total observed frequency of noun verb pair $(n,v)$, and
 $\hat{P}(n,v)$ the estimated probability of $(n,v)$, which is
 calculated as follows:
 \begin{equation} \forall n \in C_n, \forall v \in C_v,
 \hat{P}(n,v) = \frac{\hat{P}(C_n,C_v)}{|C_n \times C_v|},
 \end{equation}
\begin{equation}
 \hat{P}(C_n,C_v) = \frac{f(C_n,C_v)}{|S|}, \end{equation} where
 $f(C_n,C_v)$ denotes the observed frequency of the noun verb pairs
 belonging to cluster $(C_n,C_v)$.

 With the description length for a model defined in the above manner,
 we wish to select a model having the minimum description length and
 output it as the result of clustering. Since the model description
 length $L_{mod}$ is the same for each model, in practice we only need
 to calculate and compare $L'(M) = L_{par}(M) + L_{dat}(M)$.

 The description lengths for the data in Figure~\ref{fig:data} using
 the two models in Figure~\ref{fig:model} are shown in
 Table~\ref{tb:mdl}. (Table~\ref{tb:para} shows some values needed for
 the calculation of the description length for Model 1.) These
 calculations indicate that according to MDL, Model 1 should be
 selected over Model 2.

\begin{table*}[htb] \caption{Estimating parameters of Model 1} 
\label{tb:para}
\begin{center} \begin{tabular}{|l|c|c|c|c|} \hline
\multicolumn{1}{|c|}{$C_n \times C_v$} & $f(C_n,C_v)$ & $|C_n \times
C_v|$ & $\hat{P}(C_n,C_v)$ & $\hat{P}(n,v)$ \\ \hline
$\{{\rm wine},{\rm beer}\}\times \{{\rm eat}\}$ & $0$ & $2$ & $0$ & $0$ \\
$\{{\rm wine},{\rm beer}\}\times \{{\rm drink}\}$ & $8$ & $2$ & $0.4$ & $0.2$ \\
$\{{\rm wine},{\rm beer}\}\times \{{\rm make}\}$ & $2$ & $2$ & $0.1$ & $0.05$ \\
$\{{\rm bread},{\rm rice}\}\times \{{\rm eat}\}$ & $8$ & $2$ & $0.4$ & $0.2$ \\
$\{{\rm bread},{\rm rice}\}\times \{{\rm drink}\}$ & $0$ & $2$ & $0$ & $0$ \\
$\{{\rm bread},{\rm rice}\}\times \{{\rm make}\}$ & $2$ & $2$ & $0.1$ & $0.05$ \\ \hline
\end{tabular} \end{center} \end{table*}

\begin{table*}[htb]
\caption{Description length for the models}
\label{tb:mdl}
\begin{center}
\begin{tabular}{|l|c|} \hline
\multicolumn{2}{|c|}{Model 1} \\ \hline
$L_{par}$ & $\frac{2\times 3-1}{2}\times \log 20 =10.80$ \\
$L_{dat}$ & $-8\times \log0.2-8\times\log 0.2-2\times \log 0.05-2
\times \log 0.05=54.44$ \\
$L'$ & $10.80+54.44=65.24$ \\ \hline
\multicolumn{2}{|c|}{Model 2} \\ \hline
$L_{par}$ & $\frac{1\times 3-1}{2}\times\log 20 =4.32$ \\
$L_{dat}$ & $-8\times \log0.1-8\times\log 0.1-4 \times \log 0.05=70.44$ \\
$L'$ & $4.32+70.44=74.76$ \\ \hline
\end{tabular}
\end{center}
\end{table*}

\subsection{A Simulated Annealing-based Algorithm}
We could in principle calculate the description length for the data
using each model and select a model with the minimum description
length, if computation time were of no concern. Since the number of
probabilistic models under consideration is exponential, however, this
is not feasible in practice. We employ the `simulated annealing
technique' to deal with this problem. Figure~\ref{fig:algorithm} shows
our (divisive) algorithm for hierarchical word clustering.\footnote{As
  we noted earlier, an alternative is to employ an agglomerative
  algorithm.}
\begin{figure*}[htb]
\begin{center}
\begin{tabbing}
{\bf Algorithm:} {\bf Clustering} \\

1. Divide the noun set ${\cal N}$ into two subsets. Define a
probabilistic model consisting of the \\ 

noun partition specified by the two subsets and the entire set of verbs. \\ 

2. {\bf do} \{ \\

\tab 2.1 Randomly select one noun, remove it from the subset it
belongs to and add it to the \\

\tab other. \\ 

\tab 2.2 Calculate the description length for the two models (before
 and after the move) as $L_1$ \\ 

\tab and $L_2$, respectively. \\

\tab 2.3 Viewing the description length as the energy function for
 annealing, let $\Delta L=L_2 -L_1$.\\ 

 \tab If $\Delta L < 0$, fix the move, otherwise ascertain the
 move with probability $P =\exp(-\Delta L / T )$. \\ 

 \} {\bf while} (the description length has decreased during the
 past $10 \cdot |{\cal N}|$ trials.) \\ 

\tab Here $T$ is the annealing temperature whose initial value
is $1$ and updated to be $0.9T$ after \\ 

\tab $10 \cdot |{\cal N}|$ trials. \\ 

3. If one of the obtained subset is empty, then return the non-empty
subset, otherwise \\ 

recursively apply {\bf Clustering} on both of the two subsets. \\ 
\end{tabbing}
\caption{Simulated annealing algorithm for hierarchical word clustering}
\label{fig:algorithm}
\end{center}
\end{figure*}

\section{Advantages of Our Method}
Although there have been many methods of word clustering proposed to
date, their objectives appear different. In Table \ref{tb:comp1} and
\ref{tb:comp2} we exhibit a simple comparison between our work and
related work. Perhaps the method proposed by \cite{Pereira93} is the
most relevant in our context. In \cite{Pereira93}, they proposed a
method of `soft clustering,' in which, each word can belong to a
number of distinct classes with certain probabilities. Soft clustering
has several desirable properties. For example, word sense ambiguities
in input data can be resolved naturally. Here, we restrict our
attention on `hard clustering' (i.e., each word must belong to exactly
one class), in part because we are interested in comparing thesauruses
constructed by our method with existing hand-made thesauruses. (Note
that a hand made thesaurus is based on hard clustering.\footnote{We
  wish to investigate the possibility of employing MDL in soft
  clustering in the near future. Since MDL is a general criterion for
  statistical estimation, it can be used in other problem settings of
  word clustering. For example, recently, Stolcke \& Omohundro
  proposed to use a Bayesian model merging technique \cite{Stolcke94},
  which is similar to MDL, for the problem of word clustering in the
  context of estimating n-gram models proposed by \cite{Brown92}.})

\begin{table*}[htb]
\caption{Comparison to related work}
\label{tb:comp1}
\begin{center}
\begin{tabular}{|l|p{7cm}|l|} \hline
 & objective & co-occurrence data \\ \hline
Hindle90 & word classification & case frame data \\
Brown92 & n-gram model estimation & adjacency data \\
Pereira93 & structural and word sense disambiguation & case
frame data \\
Tokunaga95 & structural disambiguation, thesauruses for
different slots & case frame data \\
This paper & structural disambiguation, automatically constructed
thesaurus v.s. hand-made thesaurus & case frame data \\ \hline 
\end{tabular}
\end{center} 
\end{table*}

\begin{table*}[htb]
\caption{Comparison to related work}
\label{tb:comp2}
\begin{center}
\begin{tabular}{|l|l|l|} \hline
 & strategy & algorithm \\ \hline
Hindle90 & heuristics & \\
Brown92 & maximizing likelihood & agglomerative, hard
clustering \\
Pereira93 & minimizing free energy & divisive, soft clustering \\
Tokunaga95 & maximizing classification probability &
agglomerative, hard clustering \\
This paper & minimizing description length & divisive, hard clustering \\ \hline 
\end{tabular}
\end{center} 
\end{table*}

We next elaborate on the merits of our method. In statistical natural
language processing, usually the number of parameters in a
probabilistic model to be estimated is very large, and therefore such
a model is difficult to estimate with a reasonable data size that is
available in practice. (This problem is usually referred to as the
`data sparseness problem.') We could smooth the estimated
probabilities using an existing smoothing technique (e.g.,
\cite{Dagan92,Gale90}), calculate some similarity measure using the
smoothed probabilities, and then cluster words according to it. There
is no guarantee, however, that the employed smoothing method is in any
way consistent with the clustering method used subsequently. Our
method based on MDL resolves the clustering problem and the smoothing
problem in a unified fashion. (For example, the probability of the
noun verb pair (rice,make) is estimated (smoothed) to be $0.05$ in
Model 1, although the observed occurrence of it is $0$ (see
Figure~\ref{fig:data} and Figure~\ref{fig:model}).) By employing
models that embody the assumption that words belonging to the same
cluster occur with equal probability, our method achieves the
smoothing effect as a side effect of the clustering process, where the
domains of smoothing coincide with the clusters obtained by
clustering. Thus, the coarseness or fineness of clustering also
determines the degree of smoothing. All of these effects fall out
naturally as a corollary of the imperative of best possible
estimation, the original motivation behind the MDL Principle.

In our problem setting, we could alternatively employ the Maximum
Likelihood Estimator (MLE) as criterion for estimation of the best
probabilistic model, instead of MDL. MLE, as its name suggests,
selects a model which maximizes the likelihood of the data, i.e.,
$\hat{P} = \arg\max_{P} \prod_{x \in S} P(x)$. This is equivalent to
minimizing the data description length as defined in Section 3,
i.e., $\hat{P} = \arg\min_{P} \sum_{x \in S} - \log P(x)$. We can see
easily that MDL generalizes MLE, in that it also takes into account
the complexity of the model itself. In the presence of models with
varying complexity, MLE tends to overfit the data, and output a model
that is too complex and tailored to fit the specifics of the input
data. If we employ MLE as criterion for the estimation, it will result
in selecting a very fine model with many small clusters, most of which
will have probabilities estimated as zero. Thus, in contrast to
employing MDL, it will not have the effect of smoothing at all.

Purely as a strategy (criterion) of statistical estimation as well,
the superiority of MDL over MLE is supported by convincing theoretical
findings. For instance, the speed of convergence of the models
selected by MDL to the true model is known to be near optimal. (The
models selected by MDL converge to the true model approximately at the
rate of $1/s$ where $s$ is the number of parameters in the true model,
whereas for MLE the rate is $1/t$, where $t$ is the size of the
domain, or in our context, the total number of elements of ${\cal N}
\times {\cal V}$ \cite{Barron91}\cite{Yamanishi92}.) `Consistency' is
another desirable property of MDL, which is not shared by MLE. That
is, the numbers of parameters in the models selected by MDL converge
to that of the true model \cite{Rissanen84}. Both of these properties
of MDL are empirically verified in our present context, as will be
shown in the next section. In particular, we have compared the
performance of employing an MDL-based simulated annealing against that
of one based on MLE in hierarchical word clustering.

\section{Experimental Results}
We describe our experimental results in this section.

\subsection{Experiment 1: MDL v.s. MLE}
\begin{figure*}[htb]
\begin{center}
{\epsfxsize3.0in\epsfysize5.0cm\epsfbox{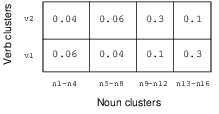}}
\caption{An artificial model}
\label{fig:model1} 
\end{center}
\end{figure*}

\begin{figure*}[htb]
\begin{center}
\begin{tabular}{lll}
\hspace{-1.0cm}
{\epsfxsize3.0in\epsfysize5.0cm\epsfbox{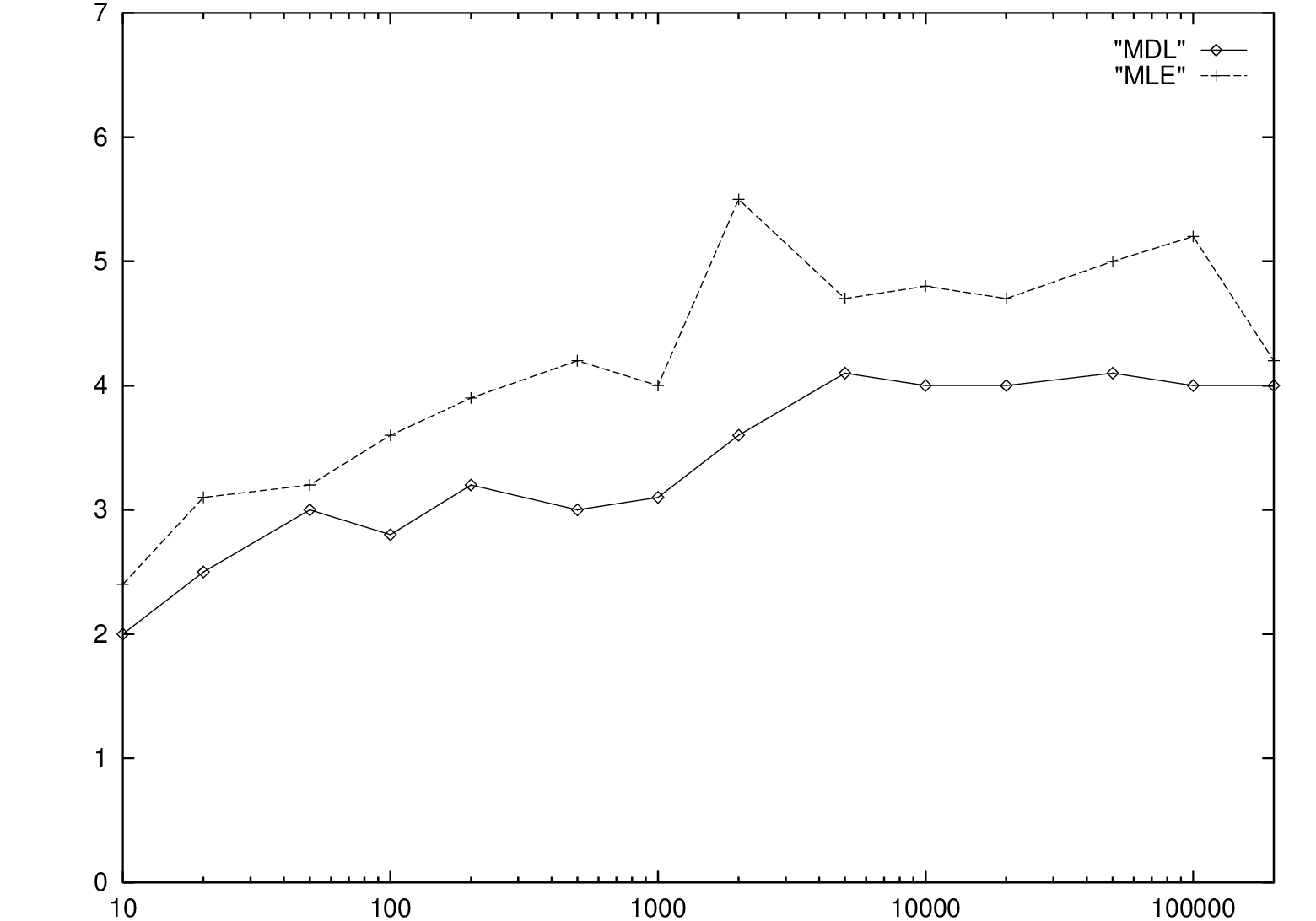}}
&
\hspace{-1.0cm}
&
{\epsfxsize3.0in\epsfysize5.0cm\epsfbox{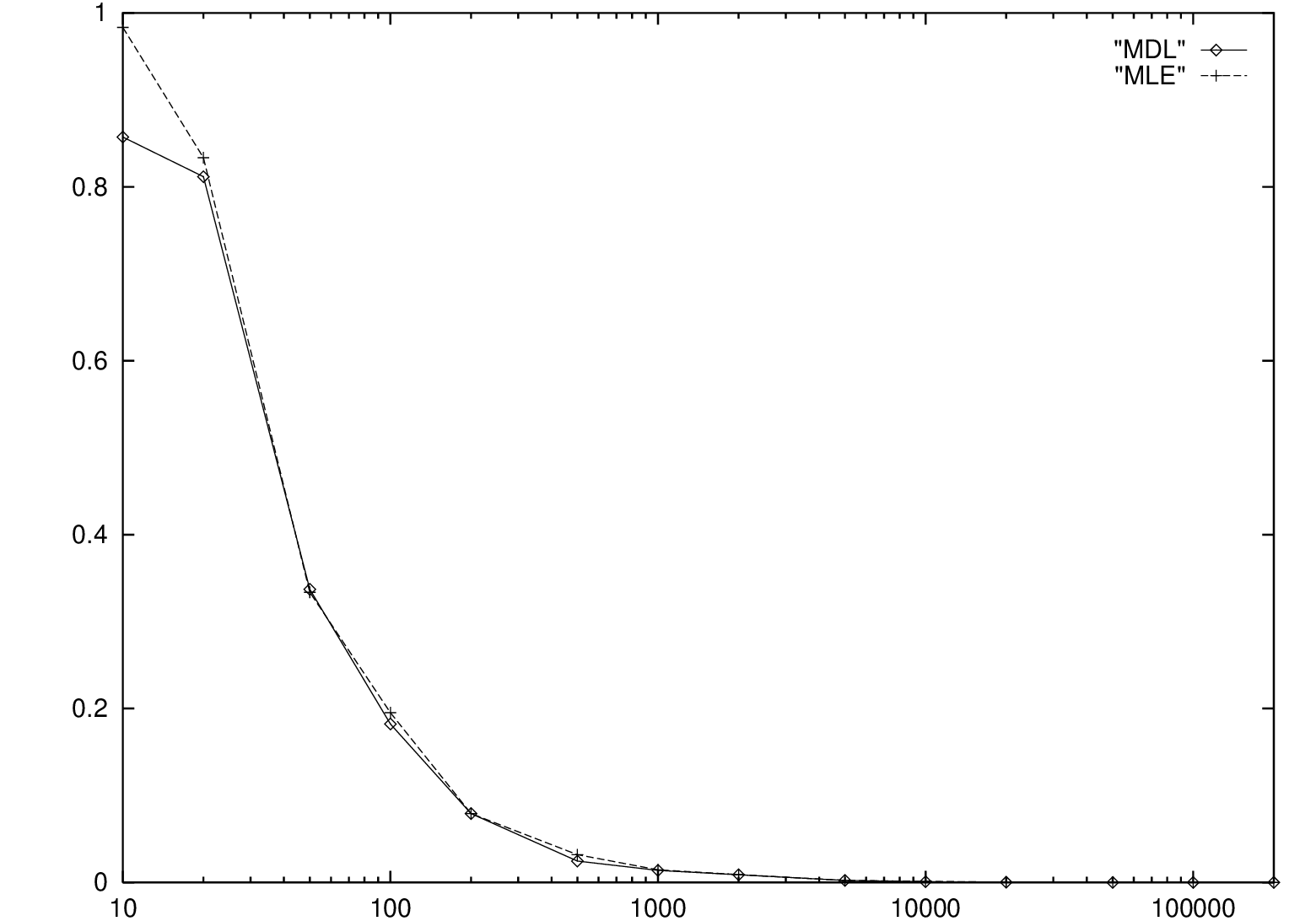}}
\end{tabular} 
\caption{(a) Number of noun clusters v.s. data size and
(b) KL distance v.s. data size}
\label{fig:model3} 
\end{center}
\end{figure*}

As described in the previous section, there are some theoretical
findings verifying that employing MDL performs better than employing
MLE in statistical estimation. We empirically test if this is the case
in our current context. We artificially constructed a {\em true} model
of word co-occurrence (see Figure~\ref{fig:model1}), and then
generated data according to its distribution. We then used the data to
estimate a model (hierarchically cluster words) by employing MDL and
MLE, respectively. (The algorithm used for MLE was the same as that
shown in Figure~\ref{fig:algorithm}, except the data description
length replaces the total description length in Step 2.)  We evaluated
the two methods in terms of the number of noun clusters and the KL
distance.\footnote{The KL distance (relative entropy), which is widely
  used in information theory and statistics, is a measure of
  `distance' between two distributions \cite{Cover91}. It is always
  non-negative and is zero iff the two distributions are identical,
  but is asymmetric and hence not a metric (the usual notion of
  distance).} Figure \ref{fig:model3}(a) plots the relation between
the number of obtained noun clusters (leaf nodes in the obtained
thesaurus tree) versus the data size, averaged over $10$ trials. (Note
that the number of noun clusters in the true model is $4$.) Figure
\ref{fig:model3}(b) plots the KL distance versus the data size, also
averaged over the same $10$ trials. The results indicate that MDL
converges to the true model faster than MLE. Also, MLE tends to select
a model that is too large (overfitting the data), while MDL tends to
select a model which is simple and yet fits the data reasonably well.
We conducted the same simulation experiments for some other models and
found the same tendencies.\footnote{The models we constructed were
  small as a model of word clustering with practical significance. We
  believe, however, that the convergence characteristics of the
  estimation methods for these models should carry over to the cases
  of estimating more practical, larger models.} We conclude that it is
better to employ MDL than MLE, as a criterion in simulated
annealing-based hierarchical word clustering.

\subsection{Experiment 2: Qualitative Evaluation}
\setlength{\unitlength}{0.5mm}
\begin{figure*}[htb]
\begin{center}
\begin{picture}(100,128)
\put(0,64){\line(1,0){20}}
\put(20,32){\line(0,1){64}}
\put(20,96){\line(1,0){20}}
\put(42,94){it he company they we i }
\put(20,32){\line(1,0){20}}
\put(40,16){\line(0,1){32}}
\put(40,48){\line(1,0){20}}
\put(60,40){\line(0,1){16}}
\put(60,56){\line(1,0){20}}
\put(82,54){one market year }
\put(60,40){\line(1,0){20}}
\put(82,38){investor official bank }
\put(40,16){\line(1,0){20}}
\put(60,8){\line(0,1){16}}
\put(60,24){\line(1,0){20}}
\put(80,20){\line(0,1){8}}
\put(80,28){\line(1,0){20}}
\put(102,26){sale loss }
\put(80,20){\line(1,0){20}}
\put(102,18){rate price }
\put(60,8){\line(1,0){20}}
\put(80,4){\line(0,1){8}}
\put(80,12){\line(1,0){20}}
\put(102,10){stock share }
\put(80,4){\line(1,0){20}}
\put(102,2){billion million }
\end{picture}
\caption{An example thesaurus}
\label{fig:thesaurus}
\end{center}
\end{figure*}
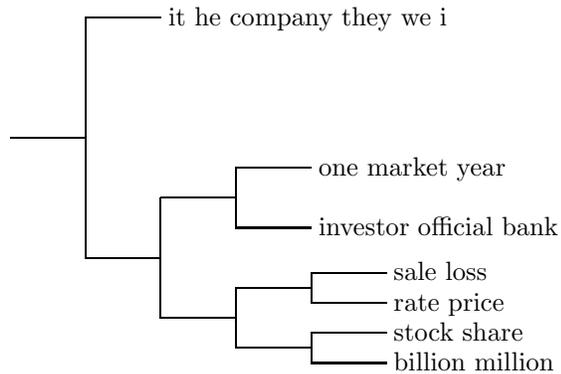

We extracted roughly $180,000$ case frames from the bracketed Wall
Street Journal (WSJ) corpus of the Penn Tree Bank \cite{Marcus93} as
co-occurrence data. We then constructed a number of thesauruses based
on these data, using our method. Figure~\ref{fig:thesaurus} shows an
example thesaurus for the $20$ most frequently observed nouns in the
data, constructed based on their appearances as subjects and objects
of roughly $2000$ verbs. The obtained thesaurus seems to agree with
human intuition to some degree. For example, `million' and `billion'
are classified in one noun cluster, and `stock' and `share' are
classified together. Not all of the noun clusters, however, seem to be
meaningful in the useful sense. This general tendency is also observed
in other example thesauruses obtained by our method. Pragmatically
speaking, however, whether the obtained thesaurus agrees with our
intuition in itself is only of secondary concern, since the main
purpose is to use the constructed thesaurus to help improve on a
disambiguation task.

\subsection{Experiment 3: Disambiguation}
We also evaluated our method by using a constructed thesaurus in
pp-attachment disambiguation experiments.

We used as training data the same $180,000$ case frames used in
Experiment 2. We also extracted as our test data $172$
$(verb,noun_1,prep,noun_2)$ patterns from the data in the same corpus,
which were {\em not} used in the training data. For the $150$ words
that appear in the position of $noun_2$, we constructed a thesaurus
based on the co-occurrences between heads and slot values of the
frames in the training data. This is because in our disambiguation
experiments we only need a thesaurus consisting of these $150$ words. 
We then applied the learning method proposed in \cite{Li95} to learn
case frame patterns using the constructed thesaurus with the same
training data as input. We formalize the case frame patterns as
conditional distributions of the form $P(Class|head,prep)$, where
$Class$ varies over the internal nodes in a certain `cut' in the
thesaurus tree.\footnote{A `cut' in a thesaurus tree defines a
  partition over the set of nouns appearing in the thesaurus.} Our
method selects the optimal cut in the thesaurus tree using the given
data in the sense of MDL, that is, a cut which is fine enough to
capture the tendency in the input data, but is coarse enough to have a
reasonably small number of parameters to estimate. It also estimates
$P(Class|head,prep)$ for each $Class$ in the cut (see \cite{Li95} for
further detail).  Table~\ref{tb:pattern} shows some example case frame
patterns obtained by this method, and Figure~\ref{fig:part} shows the
leaf nodes dominated by the internal nodes appearing in the case frame
patterns of Table~\ref{tb:pattern}.
\begin{table}[htb]
\caption{Examples of case frame patterns}
\label{tb:pattern}
\begin{center}
\begin{tabular}{|l|c|} \hline
 \multicolumn{1}{|c|}{input} & {freq.} \\ \hline
question about attitude & $1$ \\
question about corporation & $1$ \\
question about strength & $2$ \\ \hline \hline
\multicolumn{1}{|c|}{case frame pattern} & {prob.} \\ \hline
$\hat{P}(\langle {\rm strength} \rangle|{\rm question},{\rm about})$ & $0.50$ \\
$\hat{P}(\langle \#80 \rangle|{\rm question},{\rm about})$ & $0.25$ \\
$\hat{P}(\langle \#122 \rangle|{\rm question},{\rm about})$ & $0.25$ \\ \hline
\end{tabular}
\end{center}
\end{table}

\begin{figure*}[htb]
\begin{verbatim}
#80: 
ground,wake,success,network,game,rest,art,organization,plane,output,
television,benefit,letter,holder,support,nation,corporation,review,
thousand,manufacturer,margin,man,meeting,customer,agent,help

#122: 
reorganization,attitude,relief,competition,constitution
\end{verbatim}
\caption{Internal nodes and the leaf nodes they dominate}
\label{fig:part}
\end{figure*}

\begin{table}[htb]
\caption{PP-attachment disambiguation results}
\label{tb:result1}
\begin{center}
\begin{tabular}{|l|c|c|} \hline
 & coverage($\%$) & accuracy($\%$) \\ \hline
Base-Line & $100$ & $70.2$ \\
Word-Based & $19.7$ & $95.1$ \\
MDL-Thesaurus & $33.1$ & $93.0$ \\
MLE-Thesaurus & $33.7$ & $89.7$ \\
WordNet & $49.4$ & $88.2$ \\ \hline
\end{tabular}
\end{center}
\end{table}

We then conducted pp-attachment disambiguation experiments. We
compare $\hat{P}(noun_2|verb, prep)$ and $\hat{P}(noun_2| noun_1,
prep)$, which are calculated based on the case frame patterns, to
determine the attachment site of $(prep, noun_2)$.  More specifically,
if the former is larger than the latter, we attach it to $verb$; and
if the latter is larger than the former, we attach it to $noun_1$; and
otherwise (including the case in which both are $0$), we conclude that
we cannot make a decision. Table~\ref{tb:result1} shows the results of
the experiments in terms of `coverage' and `accuracy.' Here `coverage'
refers to the proportion (in percentage) of the test patterns on which
the disambiguation method can make a decision. `Base-Line' refers to
the method of always attaching $(prep,noun_2)$ to $noun_1$. 
`Word-Based,' `MLE-Thesaurus,' and `MDL-Thesaurus' respectively stand
for using word-based estimates, using a thesaurus constructed by
employing MLE, and using a thesaurus constructed by our method. Note
that the coverage of `MDL-Thesaurus' significantly outperformed that
of `Word-Based,' while basically maintaining high accuracy (though it
drops somewhat), indicating that using an automatically constructed
thesaurus can improve disambiguation results in terms of coverage.

\begin{figure*}[htb]
\begin{center}
{\epsfxsize10.0cm\epsfysize6.0cm\epsfbox{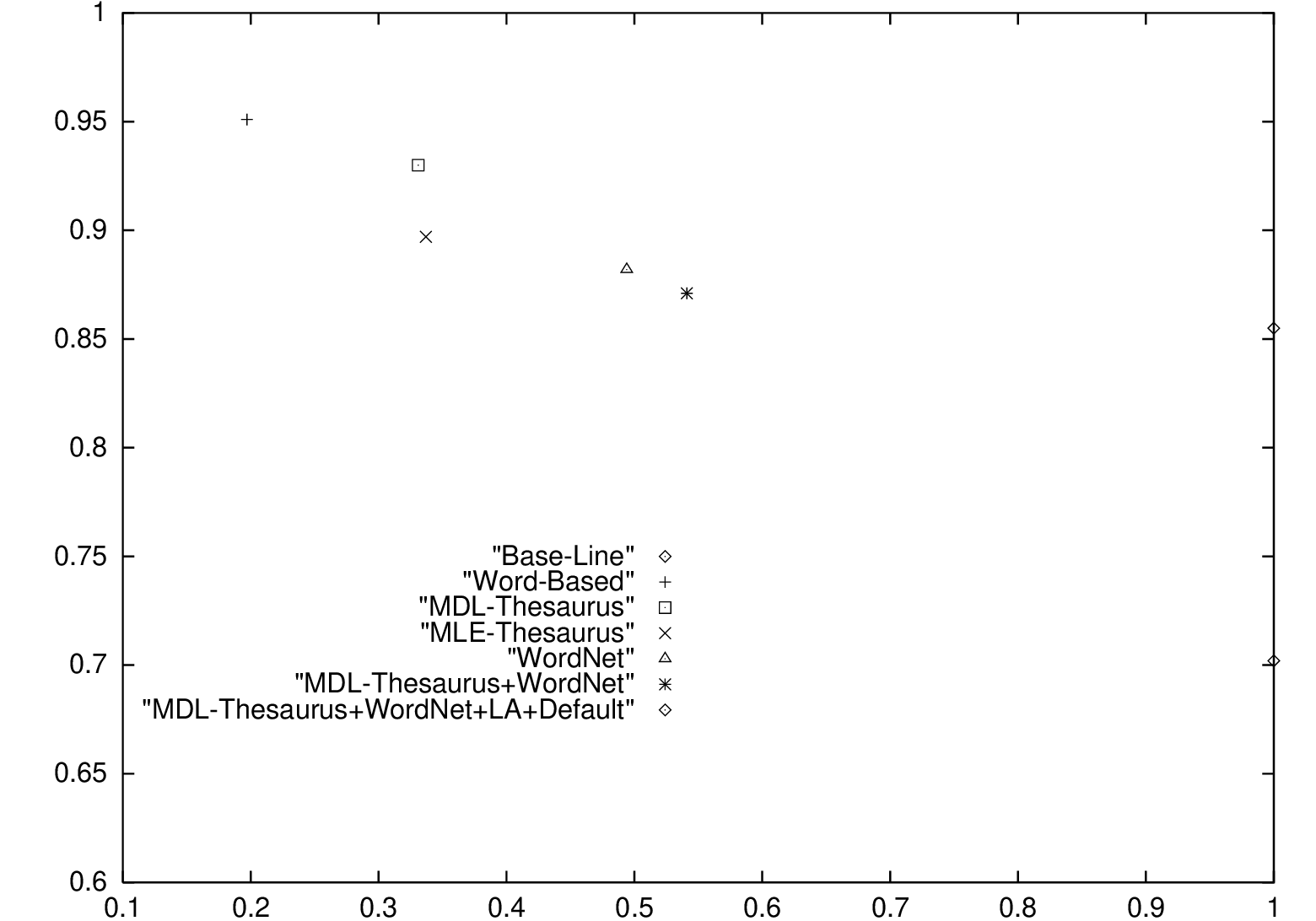}}
\caption{Accuracy v.s. coverage}
\label{fig:plot} 
\end{center}
\end{figure*}

We also tested the case of using an existing thesaurus (instead of an
automatically constructed thesaurus) to learn case frames. In
particular, we used this method with WordNet \cite{Miller93} and the
same training data, and then conducted a pp-attachment disambiguation
experiment using the obtained case frame patterns. We represent the
result of this experiment as `WordNet' in Table~\ref{tb:result1}. We
can see that in terms of coverage, WordNet outperforms MDL-Thesaurus,
but in terms of accuracy, MDL-Thesaurus outperforms WordNet. These
results can be interpreted as follows: An automatically constructed
thesaurus is more domain dependent and therefore captures the domain
dependent features better, and thus using it achieves high accuracy.
On the other hand, since training data we had available is
insufficient, its coverage is smaller than that of a hand made
thesaurus. In practice, it makes sense to combine both types of
thesauruses. That is, an automatically constructed thesaurus can be
used within its coverage, and outside its coverage, a hand made
thesaurus can be used. Given the current state of the word clustering
technique (namely, it requires data size that is usually not
available, and it tends to be computationally demanding), this
strategy is practical. We tested this strategy. More specifically, we
compare $\hat{P}(noun_2|verb,prep)$ and $\hat{P}(noun_2|noun_1,prep)$
calculated from case frame patterns obtained using an automatically
constructed thesaurus; when the two probabilities are equal, including
the case in which both are $0$, we compare the probabilities
calculated from case frame patterns obtained using WordNet. 
Table~\ref{tb:result2} represents the result of this combined method
as `MDL-Thesaurus + WordNet.' The experimental result indicates that
employing the combined method does increase the coverage of
disambiguation. We also tested `MDL-Thesaurus + WordNet + LA +
Default,' which stands for using the constructed thesaurus and WordNet
first, then the lexical association value proposed by \cite{Hindle91},
and finally the default (i.e., always attaching $prep, noun_2$ to
$noun_1$). Figure~\ref{fig:plot} shows the results. Our best
disambiguation result obtained using this last combined method
slightly improves the accuracy reported in \cite{Li95} ($84.3\%$).
\begin{table*}[htb]
\caption{PP-attachment disambiguation results}
\label{tb:result2}
\begin{center}
\begin{tabular}{|l|c|c|} \hline
 & coverage($\%$) & accuracy($\%$) \\ \hline
MDL-Thesaurus + WordNet & $54.1$ & $87.1$ \\ 
MDL-Thesaurus + WordNet + LA + Default & $100$ & $85.5$ \\ \hline
\end{tabular}
\end{center}
\end{table*}

\section{Concluding Remarks}
We have proposed a method of automatically constructing a thesaurus
(hierarchically clustering words) based on corpus data. We conclude
with the following remarks.
\begin{enumerate}
\item Our method of hierarchically clustering words based on the MDL
  Principle is theoretically sound. Our experimental results indicate
  that it is better to employ MDL than MLE as estimation criterion in
  hierarchical word clustering.
\item Using a thesaurus constructed by our method can improve
 pp-attachment disambiguation results.
\item Given the current state of the art in statistical natural language
 processing, it is best to use a combination of an automatically
 constructed thesaurus and a hand made thesaurus for disambiguation
 purpose. The disambiguation accuracy obtained this way was $85.5\%$.
\end{enumerate}

In the future, hopefully with larger training data sizes, we plan 
to construct larger thesauruses as well as to test other clustering 
algorithms.

\section*{Acknowledgement}

We thank Mr.~K.~Nakamura, Mr.~T.~Fujita, and Dr.~K.~Kobayashi of NEC
C\&C Res.~Labs.~for their constant encouragement. We thank
Dr.~K.~Yamanishi of C\&C Res.~Labs.~for his valuable comments. We
thank Ms.~Y.~Yamaguchi of NIS for her programming effort.

\end{document}